\documentclass[journal]{IEEEtran}
\IEEEoverridecommandlockouts
\usepackage{cite}
\usepackage{amsmath,amssymb,amsfonts}
\usepackage{algorithmic}
\usepackage{graphicx}
\usepackage{textcomp}
\usepackage{xcolor}
\def\BibTeX{{\rm B\kern-.05em{\sc i\kern-.025em b}\kern-.08em
    T\kern-.1667em\lower.7ex\hbox{E}\kern-.125emX}}
\begin{document}

\title{A Bibliometric Review of Neuromorphic Computing and Spiking Neural Networks}

\author{Nicholas J. Pritchard, Andreas Wicenec, Mohammed Bennamoun and Richard Dodson
\thanks{This work was supported by a Westpac Future Leaders Scholarship, an Australian Government Research Training Program Fees Offset and an Australian Government Research Training Program Stipend.}
\thanks{N.J. Pritchard, A. Wicenec and R. Dodson were with the International Centre for Radio Astronomy Research, Crawley, WA 6009, Australia. E-mail: nicholas.pritchard@icrar.org, andreas.wicenec@icrar.org, richard.dodson@icrar.org.}%
\thanks{M. Bennamoun was with is with the University of Western Australia, Crawley, WA 6009, Australia. E-mail: mohammed.bennamoun@uwa.edu.au}}
\maketitle

\begin{abstract}
Neuromorphic computing and spiking neural networks aim to leverage biological inspiration to achieve greater energy efficiency and computational power beyond traditional von Neumann architectured machines.
In particular, spiking neural networks hold the potential to advance artificial intelligence as the basis of third-generation neural networks.
Aided by developments in memristive and compute-in-memory technologies, neuromorphic computing hardware is transitioning from laboratory prototype devices to commercial chipsets; ushering in an era of low-power computing.
As a nexus of biological, computing, and material sciences, the literature surrounding these concepts is vast, varied, and somewhat distinct from artificial neural network sources. 
This article uses bibliometric analysis to survey the last 22 years of literature, seeking to establish trends in publication and citation volumes (\ref{sec:rq1}); analyze impactful authors, journals and institutions (\ref{sec:rq2}); generate an introductory reading list (\ref{sec:rq3}); survey collaborations between countries, institutes and authors (\ref{sec:rq4}), and to analyze changes in research topics over the years (\ref{sec:rq5}).
We analyze literature data from the Clarivate Web of Science using standard bibliometric methods.
By briefly introducing the most impactful literature in this field from the last two decades, we encourage AI practitioners and researchers to look beyond contemporary technologies toward a potentially spiking future of computing.  
\end{abstract}


\section{Introduction}
In recent years, contemporary artificial intelligence (AI) and deep learning methods have demonstrated rapid performance improvements \cite{schmidhuber_deep_2015, silver_mastering_2016, litjens_survey_2017}.
However, these improvements come from larger models, larger training times, larger inference times and, therefore, energy usage \cite{schwartz_green_2020}.
Artificial neural networks (ANNs), the currently dominant paradigm in AI, connect simplified models of neurons for execution on von Neumann architecture machines, forgoing the time-dependent encoding found in biological neurons for purely rate-based information encoding \cite{mcculloch_logical_1943}.
The inherent bottlenecks of scaling such devices to meet the data-intensity neural networks demand make finding improvements difficult.
Spiking neural networks take more inspiration from the underlying biological principles than artificial neural networks, and neuromorphic computing seeks to build more brain-like machines to progress beyond von Neumann bottlenecks. 
Together, the two paradigms of SNNs and neuromorphic computing are promising candidates to overcome contemporary challenges in AI and computing.
This bibliometric review summarizes the progress made in these fields during this century.
The Spiking neural network literature derives largely from biological modeling, and neuromorphic computing literature comes from a combination of computing and material sciences.
We tease out the confluence of both fields as complete end-to-end applications of neuromorphic SNNs now exist \cite{davies_advancing_2021}.
\subsection{Spiking Neural Networks (SNNs)}
Spiking Neural Networks (SNNs) are a collection of spiking neurons connected by synapses.
In biological brains, neurons communicate by sending voltage impulses, `spikes,' which modulate the activity of receiving neurons \cite{gerstner_spiking_2002}.
If a neuron reaches a sufficient level of voltage polarization, it will emit a spike and return to its resting potential.
Through this mechanism, biological brains process information with digital amplitudes and analog timings, resulting in extreme energy efficiency, as the energy demanded by a single neuron is proportional to its excitation over time; a trait neuromorphic computing aims to replicate artificially.
A broad range of neuron models exists with various goals between accurately replicating biology \cite{izhikevich_simple_2003}, allowing for efficient simulation by von Neumann machines (as with ANNs) or implementation with electronic circuitry \cite{stevens_input_1998}.

Training SNNs effectively and efficiently is still an open problem. Biological neuron behaviors change over time based on localized behavior.
Long Term Potentiation (LTP) and Long Term Depression (LTD) refer to increasing or decreasing synaptic strength with correlated or un-correlated spiking activity \cite{hebb__2002}.
This behavior inspires the Spike-Timing Dependent Plasticity (STDP) unsupervised learning rule \cite{song_competitive_2000}.
Other learning methods include training ANNs with backpropagation followed by converting the network to an SNN \cite{pfeiffer_deep_2018} and spiking variations of backpropagation \cite{bohte_error-backpropagation_2002}.
While SNN systems' performance is improving with time \cite{nunes_spiking_2022}, training methods still need to exploit the spiking time-depending encoding capabilities of SNNs fully.

Simulating spiking neuron behaviors in von Neumann machines is expensive, and simplifying time-coded neuron models into rate-coded ANNs, led to remarkable developments in AI.
The development of hardware directly implementing SNNs, neuromorphic computing seeks to add back some of this complexity in the pursuit of efficiency and effectiveness.
\subsection{Neuromorphic Computing}
Neuromorphic computing \cite{mead_neuromorphic_1990}, aims to mimic biological nervous systems as computing hardware.
Traditional von Neumann architectures exhibit a processing bottleneck when transferring instructions and data between memory and a CPU.
Co-locating memory and computing circuitry over small computational units seeks to overcome this bottleneck directly with elements communicating via addressed events \cite{boahen_communicating_1998}.
Replicating neuron dynamics comes from individual components, and digitally simulating their behaviors is also a valid option.
Exploiting the analog behaviors of sub-threshold transistors leads to exploring exotic materials.
Most recently, the advent of memristor technologies \cite{jo_nanoscale_2010}, components that retain a variable resistance proportional to their previous current, enable larger neuromorphic systems built from components implementing neuron behaviors directly.
Publicly available hardware, however, builds specialized conventional circuits to simulate SNN behavior with asynchronous execution \cite{merolla_million_2014, davies_loihi_2018}.

\subsection{Existing Reviews}
Several reviews of spiking neural networks and neuromorphic computing exist and are either extensive \cite{schuman_survey_2017} or deal with specific elements such as hardware platforms \cite{bouvier_spiking_2019}, training methods \cite{pfeiffer_deep_2018} or applications to deep learning \cite{ponulak_introduction_2011}.
Our bibliometric review differs from an existing review \cite{guan_analysis_2022} in finding important entities with citation and network analysis to point out future directions where the ANN community may contribute the most.  
\subsection{Research Questions}
A bibliometric analysis provides an overview of significant trends and an objective method to collate articles of interest.
We aim to understand:
\begin{enumerate}
    \item How publication and citation volumes are trending;
    \item Which entities are the most prolific in the field;
    \item What literature is a good starting point;
    \item Which entities collaborate the most,
    \item Whether there has been a thematic shift over the last two decades of research.
\end{enumerate}
Research into neuromorphic computing and SNNs has made significant progress recently, as have the AI and ANN communities, who are likely well-equipped to address remaining open problems.
However, there is a difference in literature lineage, and untangling the hardware developments, biological modeling, and applied AI research is challenging.
\section{Data and Methods}
In line with previous bibliometric analyses, we developed a targeted analytical strategy for literature on neuromorphic computing and SNNs.
Our methods combine standard bibliometric analysis with network measures.
\subsection{Data Collection}
We collect literature records from Clarivate's Web of Science (WoS).
Our search query selects for SNN and neuromorphic computing terms.
We consider articles published between 2000 and 2022, excluding those in early access.
We apply inclusion and exclusion criteria found in Table \ref{tab:incexcl} to filter an initial set of 28,963 articles down to 3366.
Our aggressive filtering does remove some foundational sources \cite{jo_nanoscale_2010, ohno_short-term_2011, merolla_million_2014} but ensures that our analysis focuses on SNNs and applied neuromorphic computing, forgoing works in material science.
We give current author affiliations appearing in tables.
However, their impact on institutional bibliometrics comes from their affiliation at the time of publication.
\begin{table}[htbp]
\centering
\caption{Inclusion and Exclusion Criteria}
\label{tab:incexcl}
\resizebox{\columnwidth}{!}{%
\begin{tabular}{l|l|l|l}
\hline
Spiking Neural Networks &
  Inclusion criteria &
  \begin{tabular}[c]{@{}l@{}}I1\\ I2\\ I3\\ I4\end{tabular} &
  \begin{tabular}[c]{@{}l@{}}Deep learning applications with SNNs\\ Algorithmic applications for SNNs\\ Training algorithms for SNNs\\ ANN2SNN conversion methods\end{tabular} \\ \cline{2-4} 
 &
  Exclusion criteria &
  E1 &
  Modelling of biological neurons \\ \hline
Neurmorphic Computing &
  Inclusion criteria &
  \begin{tabular}[c]{@{}l@{}}I1\\ I2\end{tabular} &
  \begin{tabular}[c]{@{}l@{}}Benchmarking hardware platforms\\ Presenting new hardware platforms\end{tabular} \\ \cline{2-4} 
 &
  Exclusion criteria &
  \begin{tabular}[c]{@{}l@{}}E1\\ E2\end{tabular} &
  \begin{tabular}[c]{@{}l@{}}Biological laboratory work\\ Resistive switching articles\end{tabular} \\ \hline
\end{tabular}}
\end{table}
\subsection{Preprocessing}
We use the original WoS data to calculate citation counts for the published literature body, institutional impact and country impact.
Bibliometrix \cite{aria_bibliometrix_2017} generates citation counts within the retained body of literature.
Bibliometric indicators and network analyses use these internal citations.
\subsection{Bibliometric Indicators}
We consider article count and citation count in all analyses.
Citation counts in country and institution impact analyses consider citations across all literature; otherwise, we use citations restricted to the retained literature.
We include additional metrics calculated for author and journal impact analyses. 

The Hirsch index (H-index) \cite{hirsch_index_2005} gives an aggregate publication and citation volume measure.
The g-index \cite{egghe_theory_2006} is the largest number such that the $g$ top-cited articles have at least $g^2$ citations. This measure allows highly successful papers to have a significant impact.
The M-index divides an author's H-index by the number of years they are active; providing an time-amortized measure of impact.
\subsection{Network Analysis}
In addition to bibliometric indicators, we employ network analysis measures to provide an alternate method to find influential works and authors. 
Our collaboration analyses employ these measures.
Network measures are calculated from citation data within the retained literature using Gephi \cite{bastian_gephi_2009}.

Page Rank \cite{page_pagerank_1999} is the original ranking algorithm developed for Google's search engine.
Entities that are linked heavily by other highly ranked entities receive a higher Page Rank.
We use a  damping factor of 0.85 and an iteration epsilon of 1e-4.

Katz Centrality \cite{katz_new_1953} measures a node's influence in a network by counting the number of nodes reached immediately and with multiple edges attenuated by a factor $\alpha$. We use an $\alpha$ of 0.5.
\subsection{Topic Clustering}
VOS viewer \cite{van_eck_software_2010} performs topic clustering over all keywords listed with each article.
Topic clustering illuminates broad trends in the content of articles when mapped over time.

\section{Results}
\subsection{Trends of publication volume (RQ1)}\label{sec:rq1}
\begin{figure}[htbp]
\centerline{\includegraphics[width=\columnwidth]{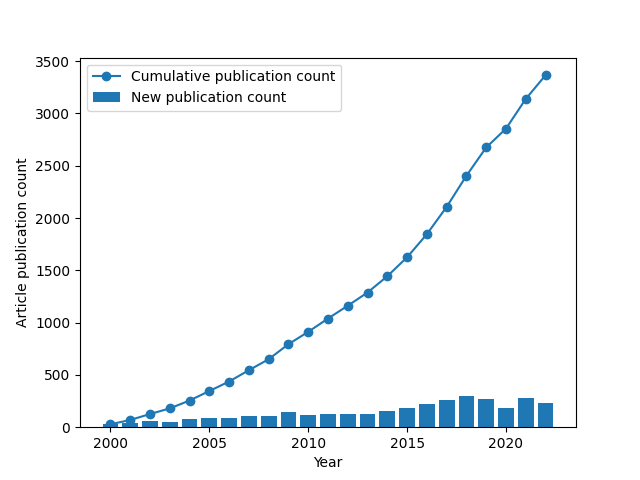}}
\caption{Article publication rate and cumulative count.}
\label{fig:pubcount}
\end{figure}

\begin{figure}[htbp]
\centerline{\includegraphics[width=\columnwidth]{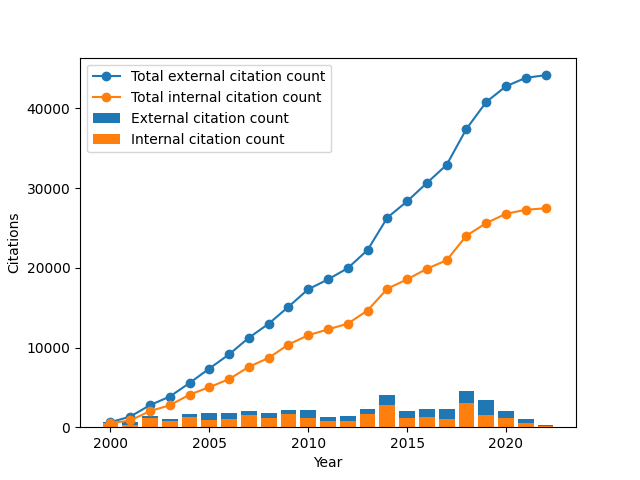}}
\caption{Total citation rate and cumulative count.}
\label{fig:combinedcite}
\end{figure}

\begin{figure}[htbp]
    \centering
    \includegraphics[width=\columnwidth]{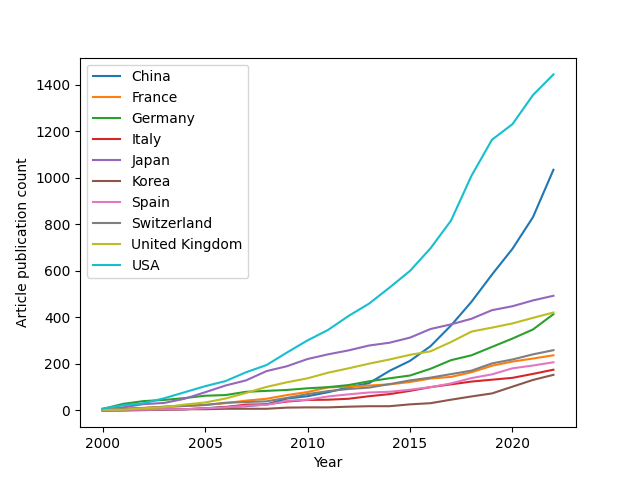}
    \caption{Article publication rate over time per country.}
    \label{fig:country-pubcount}
\end{figure}

Fig. \ref{fig:pubcount} displays the rate of new article publication and cumulative publication count.
The publication rate is trending positively.
Fig. \ref{fig:combinedcite} displays the rate and cumulative count of citations within the queried literature records and externally among all literature.
Both the internal and external citation rate increases sharply after 2012 and in 2018 with an outsized impact external to the queried literature.
Fig. \ref{fig:country-pubcount} shows the number of articles published per year by the top 10 most productive countries.
We observe a general upwards trend with the USA and China lead by a wide margin.

The publication production rate and citation rates are increasing in this field.
The advent of commercially available neuromorphic computing hardware will hopefully further accelerate this trend.
\subsection{Analyses of Impactful Authors, Journals, Institutions and Countries (RQ2)}\label{sec:rq2}
In total, 6676 authors from 78 countries from 1852 institutions contributed to the 3366 articles published between 1249 outlets. 
Table \ref{tab:rq2Authors} lists the 15 most prolific authors ordered by H-index.
Nine of the ten most prolific are affiliated outside the United States, and the H-index, g-index and i10 metrics generally agree with this ordering.

\begin{table*}[htbp]
\centering
\caption{Top Authors Ranked by H-Index}
\label{tab:rq2Authors}
\begin{tabular}{lllllll}
\hline
Name               & Institution                                              & H-Index & G-Index & M-Index & TC   & NP \\
\hline
Nikola Kasabov    & Auckland University of Technology                        & 22      & 42      & 1.1     & 1956 & 88 \\
Steve Furber      & University of Manchester                                 & 17      & 42      & 0.85    & 1845 & 61 \\
Giacomo Indiveri  & University of Zurich and ETH Zurich                      & 16      & 45      & 0.696   & 2088 & 55 \\
Wolfgang Maass    & Graz University of Technology                            & 13      & 22      & 0.542   & 1103 & 22 \\
Arindam Basu      & University of Canterbury                                 & 12      & 21      & 0.706   & 474  & 21 \\
Johannes Schemmel & Heidelberg University                                    & 12      & 25      & 0.667   & 786  & 25 \\
Ammar Belatreche  & Northumbria University                                   & 11      & 22      & 0.524   & 496  & 24 \\
Peng Li           & University of California, Santa Barbara                  & 11      & 19      & 0.846   & 390  & 29 \\
Shih-Chii Liu     & University of Zurich and ETH Zurich                      & 11      & 27      & 0.478   & 770  & 32 \\
Karlheinz Meier   & Heidelberg University                                    & 11      & 22      & 0.611   & 759  & 22 \\
Luis Plana        & Barcelona Supercomputing Center                          & 11      & 19      & 0.688   & 1487 & 19 \\
Andre van Schaik   & Western Sydney University                                & 11      & 23      & 0.478   & 572  & 25 \\
Gert Cauwenberghs & University of California, San Diego                      & 10      & 19      & 0.435   & 540  & 19 \\
Hong Qu           & University of Electronic Science and Technology of China & 10      & 18      & 0.667   & 335  & 19 \\
\hline
\multicolumn{7}{l}{Abbreviations: NP: Number of Publications; TC: Total Citations. }
\end{tabular}
\end{table*}

Table \ref{tab:rq2Journals} lists the 10 most influential journals ordered by H-index.
The top five journals are specific to the field of computational neuroscience.
Notably, while indexed separately, IEEE Transactions On Neural Networks and Learning Systems succeeds IEEE Transactions On Neural Networks.

\begin{table*}[htbp]
\centering
\caption{Top Journals Ranked by H-Index}
\label{tab:rq2Journals}
\begin{tabular}{lllll}
\hline
Name                                                       & H-Index & G-Index & TC   & NP  \\
\hline
Neural Computation                                         & 32      & 66      & 4823 & 141 \\
Neural Networks                                            & 30      & 56      & 3536 & 117 \\
Neurocomputing                                             & 25      & 47      & 2924 & 168 \\
IEEE Transactions On Neural Networks and Learning Systems  & 24      & 39      & 1723 & 67  \\
IEEE Transactions On Neural Networks                       & 19      & 30      & 2057 & 30  \\
International Journal Of Neural Systems                    & 17      & 24      & 740  & 24  \\
Scientific Reports                                         & 17      & 28      & 877  & 48  \\
IEEE Transactions On Circuits And Systems I-Regular Papers & 16      & 28      & 809  & 30  \\
IEEE Transactions On Biomedical Circuits And Systems       & 14      & 27      & 754  & 27  \\
Nature Communications                                      & 13      & 21      & 633  & 21 \\
\hline 
\multicolumn{5}{l}{Abbreviations: Same as Table \ref{tab:rq2Authors}.}
\end{tabular}
\end{table*}

Table \ref{tab:rq2Institutions} lists the most influential institutions by citation count. Four of the top ten are in the United States, seemingly paradoxical considering that the majority of the most prolific individuals are associated outside of North America. 
Interesting enough, two companies (one defunct) reach into the top ten; suggesting a recent commercial interest in neuromorphic technology.

\begin{table}[htbp]
\centering
\caption{Top Institutes by Total Citations.}
\label{tab:rq2Institutions}
\begin{tabular}{lll}
\hline
Institution                           & TC   & NP \\
\hline
University of Zurich                  & 3173 & 98 \\
Auckland University of Technology     & 2188 & 91 \\
Intel Corporation                     & 1972 & 20 \\
University of Manchester              & 1944 & 69 \\
Swiss Federal Institute of Technology & 1716 & 66 \\
Princeton University                  & 1455 & 14 \\
Stanford University                   & 1451 & 22 \\
Graz University of Technology         & 1257 & 26 \\
ETH Zurich                            & 1202 & 19 \\
Reduced Energy Microsystems           & 1143 & 1  \\
Ulster University                     & 870  & 58 \\
Leiden University                     & 768  & 3  \\
University of California San Diego    & 749  & 34 \\
University of Southampton             & 709  & 15 \\
Ecole Normale Super                   & 669  & 9  \\
\hline
\multicolumn{3}{l}{Abbreviations: Same as Table \ref{tab:rq2Authors}.}
\end{tabular}
\end{table}

Table \ref{tab:rq2Country} lists the most influential countries by citation count.
The results are commensurate with general trends in academia, and it is noteworthy that the US and China published approximately 40\% of the total publications combined.
Considering average citations per publication, however, reveals smaller countries like Switzerland make outsized contributions to the field.

\begin{table}[htbp]
\centering
\caption{Top Countries/Regions Ranked by Citation Count.}
\label{tab:rq2Country}
\begin{tabular}{lll}
\hline
Country                  & TC    & NP  \\
\hline
United States of America & 15083 & 838 \\
China                    & 5467  & 522 \\
Switzerland              & 4805  & 172 \\
United Kingdom           & 4789  & 300 \\
Germany                  & 4027  & 238 \\
France                   & 2370  & 163 \\
New Zealand              & 2331  & 106 \\
Japan                    & 2052  & 340 \\
Austria                  & 1551  & 44  \\
Netherlands              & 1242  & 43  \\
Spain                    & 1161  & 154 \\
Singapore                & 1125  & 83  \\
Italy                    & 1098  & 109 \\
Ireland                  & 1039  & 84  \\
Australia                & 1002  & 81  \\
\hline
\multicolumn{3}{l}{Abbreviations: Same as Table \ref{tab:rq2Authors}.}
\end{tabular}
\end{table}

\subsection{Essential Reading (RQ3)}\label{sec:rq3}
Table \ref{tab:rq3Works} lists the top 15 items ranked by total citation count.
A mix of training algorithms \cite{bohte_error-backpropagation_2002, ponulak_supervised_2010, kasabov_dynamic_2013, ghosh-dastidar_new_2009, mohemmed_span_2012, bohte_unsupervised_2002, legenstein_what_2005, wade_swat_2010}, full end-to-end systems \cite{davies_loihi_2018, benjamin_neurogrid_2014, indiveri_vlsi_2006, furber_overview_2013, furber_spinnaker_2014}, application to neuroscience investigation \cite{kasabov_neucube_2014}, and extension to reinforcement learning \cite{florian_reinforcement_2007} make up the top works.
\begin{table}[htbp]
\centering
\caption{Top Papers Ranked by Local Citation Count.}
\label{tab:rq3Works}
\begin{tabular}{lll}
\hline
Document              & LC  & TC   \\
\hline
\cite{bohte_error-backpropagation_2002} Bohte (2002)          & 291 & 587  \\
\cite{davies_loihi_2018} Davies (2018)         & 260 & 1143 \\
\cite{benjamin_neurogrid_2014} Benjamin (2014)       & 169 & 639  \\
\cite{indiveri_vlsi_2006} Indiveri (2006)       & 152 & 660  \\
\cite{ponulak_supervised_2010} Ponulak (2010)        & 148 & 325  \\
\cite{furber_spinnaker_2014} Furber (2014)         & 134 & 605  \\
\cite{furber_overview_2013} Furber (2013)         & 100 & 362  \\
\cite{kasabov_dynamic_2013} Kasabov (2013)        & 99  & 205  \\
\cite{kasabov_neucube_2014} Kasabov (2014)        & 92  & 217  \\
\cite{ghosh-dastidar_new_2009} Ghosh-Dastidar (2009) & 88  & 332  \\
\cite{bohte_unsupervised_2002} Bohte (2002)          & 87  & 156  \\
\cite{mohemmed_span_2012} Mohemmed (2012)       & 74  & 172  \\
\cite{legenstein_what_2005} Legenstein (2005)     & 63  & 149  \\
\cite{wade_swat_2010} Wade (2010)           & 62  & 117  \\
\cite{florian_reinforcement_2007} Florian (2007)        & 61  & 182  \\
\hline
\multicolumn{3}{l}{Abbreviations: LC: Local citations; TC: Total Citations.}
\end{tabular}
\end{table}

Table \ref{tab:rq3PageRank} lists the top 15 items ranked by Page Rank.
While several items appear also in Table \ref{tab:rq3Works}, those that do not, include system developments and demonstrations \cite{park_hierarchical_2017, bofill-i-petit_synchrony_2004, bamford_large_2010}, exploration of technical performance through simulation \cite{richmond_democratic_2011, morrison_advancing_2005}, training algorithms \cite{tavanaei_bp-stdp_2019} and neuroscience exploration \cite{dong_estimating_2011}.
\begin{table}[htbp]
\centering
\caption{Top Papers Ranked by Page Rank}
\label{tab:rq3PageRank}
\begin{tabular}{lll}
\hline
Document              & Page Rank & TC \\
\hline
\cite{ponulak_supervised_2010} Ponulak (2010)        & 2.29E-02  & 325  \\
\cite{davies_loihi_2018} Davies (2018)         & 1.55E-02  & 1143 \\
\cite{kasabov_dynamic_2013} Kasabov (2013)        & 1.28E-02  & 205  \\
\cite{park_hierarchical_2017} Park (2017)           & 1.15E-02  & 52   \\
\cite{florian_reinforcement_2007} Florian (2007)        & 1.11E-02  & 182  \\
\cite{richmond_democratic_2011} Richmond (2011)       & 1.07E-02  & 10   \\
\cite{tavanaei_bp-stdp_2019} Tavanaei (2019)       & 9.79E-03  & 65   \\
\cite{bohte_error-backpropagation_2002} Bohte (2002)         & 9.40E-03  & 587  \\
\cite{indiveri_vlsi_2006} Indiveri (2006)       & 9.37E-03  & 660  \\
\cite{bofill-i-petit_synchrony_2004} Bofill-I-Petit (2004) & 8.49E-03  & 104  \\
\cite{legenstein_what_2005} Legenstein (2005)     & 7.91E-03  & 149  \\
\cite{benjamin_neurogrid_2014} Benjamin (2014)       & 6.79E-03  & 639  \\
\cite{morrison_advancing_2005} Morrison (2005)       & 6.39E-03  & 120  \\
\cite{dong_estimating_2011} Dong (2011)           & 6.26E-03  & 13   \\
\cite{bamford_large_2010} Bamford (2010)        & 5.91E-03  & 15   \\
\hline
\multicolumn{3}{l}{Abbreviations: Same as Table \ref{tab:rq2Authors}.}
\end{tabular}
\end{table}

We supplement our metric-derived reading lists with a hand-crafted list for those with a computer science background.
We recommend \cite{laurence_f_abbott_theoretical_2001, trappenberg_fundamentals_2010} for textbook resources.
Foundational works include \cite{mead_neuromorphic_1990}, for the foundation of the field and \cite{jo_nanoscale_2010} for background in memristor technologies.
Some notable, introductory hardware systems include \cite{indiveri_vlsi_2006} for an analog VLSI implementation, IBM and related developments \cite{benjamin_neurogrid_2014, merolla_million_2014} and Intel's efforts \cite{davies_loihi_2018, davies_advancing_2021}.
Good overviews of the field include \cite{yamazaki_spiking_2022} for training algorithms, \cite{roy_towards_2019} for a historical account focusing on hardware-software co-design and \cite{rajendran_low-power_2019} which reviews contemporary neuromorphic chip architectures.
These papers and texts provide a suitable collection to start a journey in neuromorphic computing, covering a wide range of topics at both the leading edge and foundational material.
\subsection{Collaboration Analysis (RQ4)}\label{sec:rq4}
In addition to ranking works and entities by citations, network analysis of collaboration patterns illuminates how prominent authors, institutes and countries achieve their success.
Table \ref{tab:rq4Authors} lists the top ranked 15 authors by collaborative Katz Centrality acting on their co-authorship links.
Highly-active younger academics dominate these metrics, which bodes well for the future of the field.
\begin{table}[htbp]
\centering
\caption{Top Authors Ranked by Collaboration Katz Centrality}
\label{tab:rq4Authors}
\begin{tabular}{lll}
\hline
Name                 & Katz Centrality & Page Rank \\
\hline
Stefan Schliebs      & 205.60          & 1.19E-03  \\
Runchun Wang         & 176.67          & 9.87E-04  \\
Steve Temple         & 138.05          & 2.94E-03  \\
Sandeep Pande        & 105.30          & 7.97E-04  \\
Ammar Mohemmed       & 79.20           & 8.41E-04  \\
Neelava Sengupta     & 73.18           & 1.58E-03  \\
Fearghal Morgan      & 71.69           & 1.01E-03  \\
Luis A. Plana        & 71.53           & 2.72E-03  \\
Andre van Schaik     & 64.99           & 1.55E-03  \\
Bernhard Vogginger   & 63.53           & 2.32E-03  \\
Jie Yang             & 60.81           & 2.05E-03  \\
Malu Zhang           & 50.43           & 1.93E-03  \\
Nuttapod Nuntalid    & 48.81           & 8.20E-04  \\
Amirreza Yousefzadah & 47.07           & 9.89E-04  \\
Johannes Schemmel    & 45.80           & 2.08E-03  \\
\hline
\end{tabular}
\end{table}
Table \ref{tab:rq4Institute} ranks the top institutes by Katz Centrality. Here we see strong teams from outside the United States be the most collaborative. 
\begin{table}[htbp]
\centering
\caption{Top Institutions Ranked by Collaboration Katz Centrality}
\label{tab:rq4Institute}
\begin{tabular}{lll}
\hline
Institution                                              & Katz Centrality & Page Rank \\
\hline
University of Zurich                                     & 488.32          & 8.69E-03  \\
University of Southampton                                & 241.98          & 1.47E-03  \\
University of York                                       & 204.23          & 9.73E-04  \\
Xi'an Jiaotong University                                & 164.88          & 2.45E-03  \\
University of Waterloo                                   & 133.12          & 3.23E-03  \\
Xihua University                                         & 121.75          & 1.77E-03  \\
University of Seville                                    & 112.30          & 4.22E-03  \\
\begin{tabular}[c]{@{}l@{}}University of Electronic Science and \\ Technology of China\end{tabular} & 93.96 & 4.10E-03 \\
Zhejiang University of Technology                        & 85.85           & 6.23E-04  \\
Yahoo! Research                                          & 84.30           & 6.72E-04  \\
\begin{tabular}[c]{@{}l@{}}Xi'an University of \\ Science and Technology\end{tabular}               & 83.44 & 3.18E-04 \\
Zhejiang University                                      & 83.19           & 2.43E-03  \\
Ulster University                                        & 82.76           & 4.01E-03  \\
University of California, Santa Barbara                  & 70.67           & 1.89E-03  \\
University of Manchester                                 & 59.41           & 5.15E-03  \\
\hline
\end{tabular}
\end{table}
Table \ref{tab:rq4Country} list the top 15 countries ranked by collaborative Katz Centrality.
The presence of several Chinese institutions in Table \ref{tab:rq4Institute} is commensurate with China ranking highly.
Sparing appearance from American institutes is surprising but understandable given the strong domestic research ecosystem in the United States.
\begin{table}[htbp]
\centering
\caption{Top Countries / Regions Ranked by Collaboration Katz Centrality}
\label{tab:rq4Country}
\begin{tabular}{lll}
\hline
Name                     & Katz Centrality & Page Rank \\
\hline
United States of America & 28202694.51     & 5.11E-02  \\
Switzerland              & 2805929.29      & 3.26E-02  \\
Ukraine                  & 1420345.52      & 6.66E-03  \\
Spain                    & 664085.03       & 3.89E-02  \\
Singapore                & 642014.50       & 1.70E-02  \\
Turkey                   & 285873.19       & 9.66E-03  \\
Scotland                 & 236109.80       & 1.79E-02  \\
South Africa             & 80377.56        & 9.77E-03  \\
Saudi Arabia             & 72332.36        & 1.67E-02  \\
China                    & 68438.26        & 4.24E-02  \\
Portugal                 & 51678.50        & 1.35E-02  \\
Tunisia                  & 36167.18        & 3.54E-03  \\
Sweden                   & 34688.23        & 1.31E-02  \\
Poland                   & 34320.27        & 1.69E-02  \\
Taiwan                   & 17113.93        & 5.96E-03  \\
\hline
\end{tabular}
\end{table}
\subsection{Research Topics and Clustering (RQ5)}\label{sec:rq5}
\begin{figure*}[htbp]
    \centering
    \includegraphics[width=\textwidth]{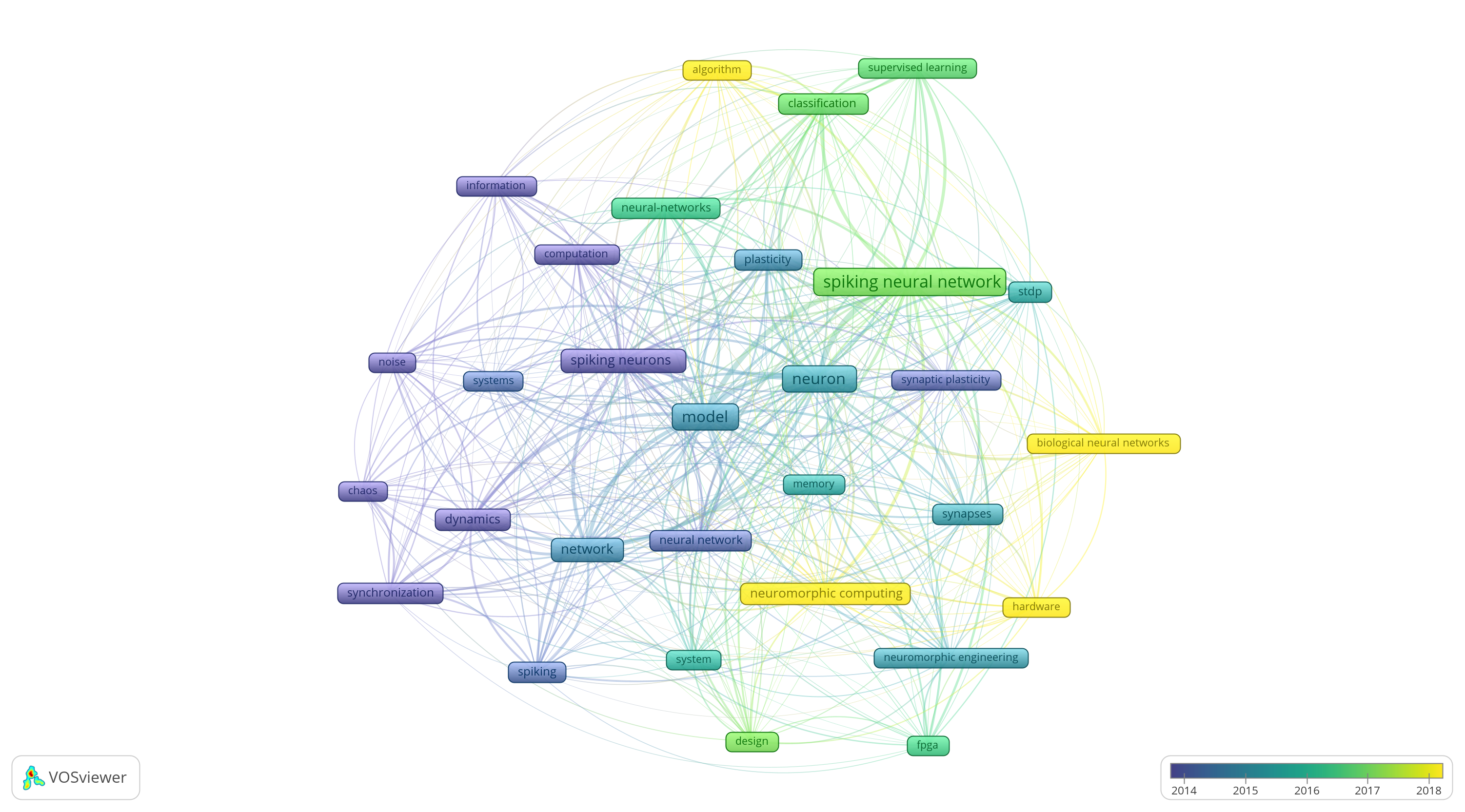}
    \caption{Terms clustering visualization from VOS Viewer. Fundamental concepts find use across all times, and there is a trend towards hardware-related terms over time; reflecting the development of neuromorphic hardware platforms.}
    \label{fig:rq5Terms}
\end{figure*}

Figure \ref{fig:rq5Terms} contains a network of keywords related by co-occurrence across all items and color coded by publication date.
While concepts such as `spiking neurons', `neurons' and `neural networks' are present across several years; there is a distinct trend towards hardware concepts like `hardware' and `neuromorphic computing'.
The transition, broadly speaking, from theory to implementation bodes well for future applications of SNNs and neuromorphic computing.

\section{Discussion}
We discuss a few significant developments from the last 22 years of literature and suggest critical areas for future development.
\subsection{Neuromorphic Hardware Platforms}
The literature places great emphasis on the advent of nascent neuromorphic computing platforms.
Built upon vast swathes of research, these platforms and their introductory papers \cite{merolla_million_2014, davies_loihi_2018} tie together significant developments from the material science required to construct memristive devices, algorithmic work to make the hardware sing and novel applications to demonstrative problems.
Generally, there is a shift from highly specialized lab-focused endeavors aiming at biological modeling to commercial attempts with more general-purpose uses in mind. 
\subsection{Training SNNs}
Core to applying nascent neuromorphic hardware designed to implement SNNs is their training. Algorithms in this arena derive from biological observations or adaptation of successful approaches for ANNs.
Innovative approaches outside these two approaches are likely to yield exciting results, inform how to use SNNs in practice, and illuminate our general understanding of intelligence \cite{yamazaki_spiking_2022}.
\subsection{Future Research Directions}
To this end, we present some suggestions for future research to advance the field of neuromorphic computing and SNNs.
Training SNNs, especially on nascent hardware, is an open problem with great potential. While SNN performance on ANN benchmarks is improving, we suggest finding novel benchmarks best suited to illustrate the unique capabilities SNNs offer and furthering research into online and real-time SNN learning.
Developing neuromorphic chips is the keystone development of the last two decades, and continuing research into novel materials, architectures and memristive devices will complement finding applications for neuromorphic hardware.
Specifically, scaling neuromorphic systems and finding application-specific designs will test the limits of SNNs.
ANNs found widespread adoption and investigation with the advent of sophistical tooling, lowering the barrier of entry to testing novel ideas; continued development into SNN tooling is essential to progressing the field.
Moreover, integrating SNN technologies into existing AI paradigms and investigating the tradeoff between ANNs and energy-efficient SNNs will search out applications in several domains, such as natural language processing and computer vision.
Building upon existing SNN frameworks and attempts to use them will inform training algorithms and future hardware platforms.
\section{Conclusion}
This article evaluates the development of scientific research into neuromorphic computing and SNNs in this century with bibliometric and network analysis.
SNNs and neuromorphic computing have progressed from biological modeling and material science to constructing end-to-end systems realizing SNNs in digital and analog hardware.
A wide array of teams distributed globally make up the field, with the most impactful individuals residing outside of the United States. At the same time, the US retains the most impact on aggregate.
The most significant journal outlets are specific to the field of neuroscience and we provide a guided list of impactful literature for those from a background in traditional machine learning and ANN research.
We remark on future directions in the field, largely oriented toward accelerating the evaluation of neuromorphic hardware, in comparison to ANNs and to inform future hardware designs.
There is a vast world of problems to solve, and the ANN community is well-equipped to make progress.  
\bibliography{Bibliometric-SNN}{}
\bibliographystyle{IEEEtran}
\end{document}